\documentclass[10pt,twocolumn,letterpaper]{article}

\usepackage{iccv}
\usepackage{times}
\usepackage{epsfig}
\usepackage{graphicx}
\usepackage{amsmath}
\usepackage{amssymb}

\usepackage{subcaption}
\usepackage[dvipsnames,table,xcdraw]{xcolor}
\usepackage{colortbl}
\usepackage{stmaryrd}
\usepackage{multirow}
\usepackage{booktabs}
\usepackage{pgfplots}
\usepackage{xspace}
\usepackage{subcaption}
\usepackage[export]{adjustbox}

\usepackage[pagebackref=true,breaklinks=true,letterpaper=true,colorlinks,bookmarks=false]{hyperref}

\iccvfinalcopy 

\def\iccvPaperID{10} 

\ificcvfinal\fi

\begin{document}

\title{BluNF: Blueprint Neural Field}

\author{Robin Courant$^1$\thanks{Equal contribution.} \hspace{30pt} Xi Wang$^1$\footnotemark[1] \hspace{30pt} Marc Christie$^2$ \hspace{30pt} Vicky Kalogeiton$^1$\\
{\normalsize $^1$LIX, Ecole Polytechnique, IP Paris} ~~~~
{\normalsize $^2$Inria, IRISA, CNRS, Univ. Rennes}\\
}

\maketitle
\ificcvfinal\thispagestyle{empty}\fi

\begin{abstract}

Neural Radiance Fields (NeRFs) have revolutionized scene novel view synthesis, offering visually realistic, precise, and robust implicit reconstructions. While recent approaches enable NeRF editing, such as object removal, 3D shape modification, or material property manipulation, the manual annotation prior to such edits makes the process tedious. 
Additionally, traditional 2D interaction tools lack an accurate sense of 3D space, preventing precise manipulation and editing of scenes.
In this paper, we introduce a novel approach, called \textbf{Bl}ueprint \textbf{N}eural \textbf{F}ield (BluNF), to address these editing issues. 
BluNF provides a robust and user-friendly 2D blueprint, enabling intuitive scene editing. By leveraging implicit neural representation, BluNF constructs a blueprint of a scene using prior semantic and depth information.
The generated blueprint allows effortless editing and manipulation of NeRF representations. We demonstrate BluNF's editability through an intuitive click-and-change mechanism, enabling 3D manipulations, such as masking, appearance modification, and object removal.
Our approach significantly contributes to visual content creation, paving the way for further research in this area. 

\end{abstract}


\section{Introduction}
\begin{figure}[hbt!]
  \centering
  \includegraphics[width=\columnwidth]{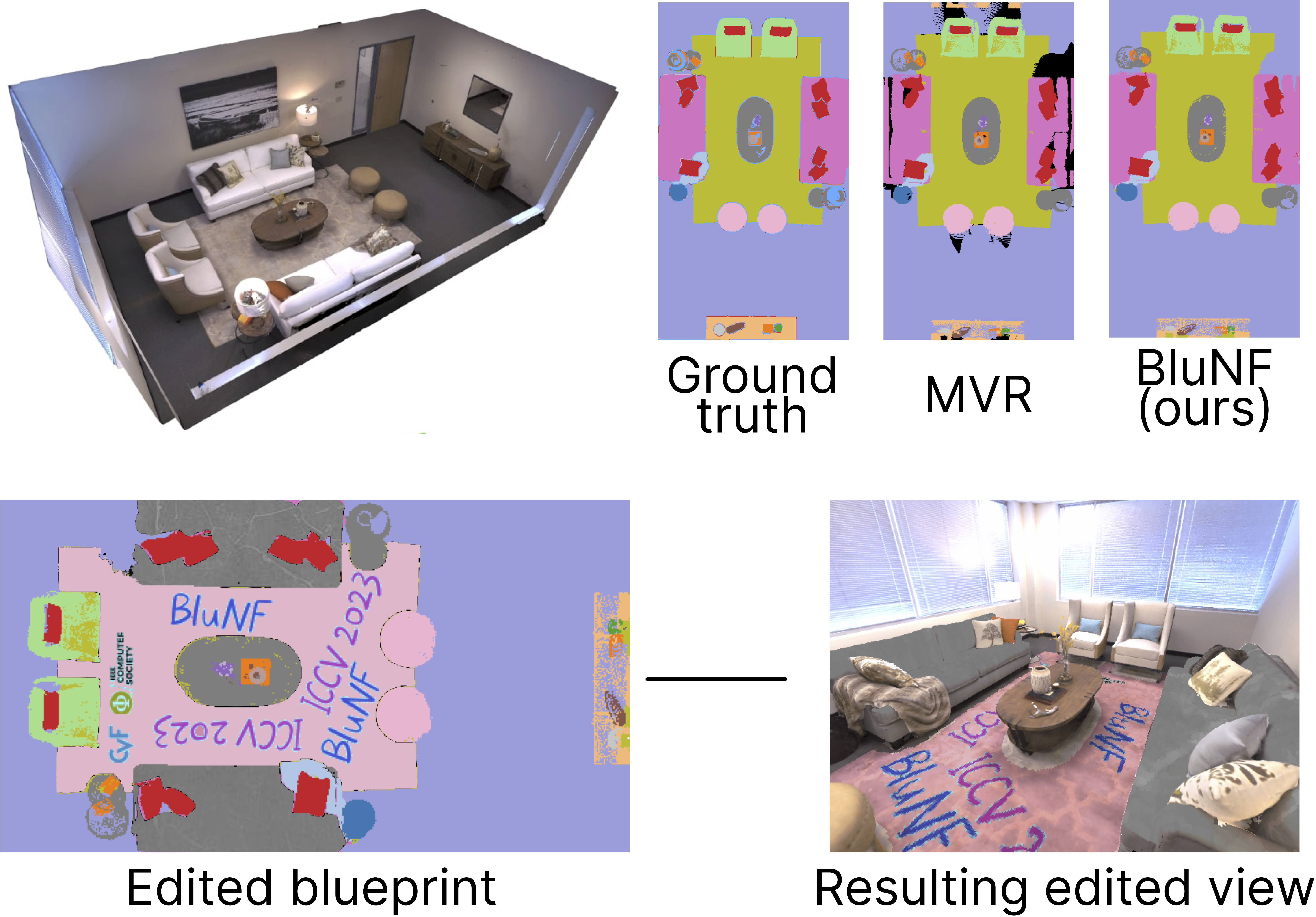}
  \caption{
  Generating a blueprint from multiple views of a scene can be challenging. (top) Traditional multiple-view reconstruction methods (MVR) depend on the diversity of views, and suffer from occlusions (black areas in MVR blueprint). Instead, our proposed BluNF addresses these by leveraging an implicit neural representation of the scene. (bottom) BluNF associated with NeRF enables scene editing, such as changes in appearance or texture. 
  }
  \label{fig:teaser}
\end{figure}

The demand for realistic and taylored 3D scenes is increasing for various applications, ranging from artistic purposes like movie and video game creation to more practical uses in architecture and design. 
While traditional methods, like photogrammetry and light-field encoding, have facilitated the reconstruction of 3D scenes from multiple viewpoints~\cite{Schonberger_2016_CVPR, levoy1996light}, recent advancements in deep learning, such as Neural Radiance Fields (NeRF)~\cite{mildenhall2021nerf,barron2021mip, barron2022mip, turki2022mega} have shown impressive capabilities in capturing and rendering realistic scenes.  However, editing and manipulating NeRF representations remains challenging.

Existing works propose various NeRF editing techniques, such as scene composition using multiple NeRFs~\cite{yang2021learning}, manipulation of rendering properties~\cite{kuang2022palettenerf}, deformations through mesh-to-NeRF mappings~\cite{yuan2022nerf}, and object removal using user-provided masks~\cite{Weder2023Removing}. 
However, selecting and editing NeRF views present specific challenges, particularly when relying on traditional 2D interaction tools that lack a true sense of 3D space (e.g. 2D screens, keyboards and mice).
Especially when considering the needs of designers/users who often rely on simplified or partial views of scenes for higher-level semantic manipulation.
Blueprints, commonly used in architectural designs, establish a clear link between the semantic representation of entities and their underlying geometric structure, providing a foundation for constructing real scene layouts. 
Beyond architectural applications, blueprints find utility in diverse computer vision and graphics applications, including motion planning~\cite{ichter2020learned}, navigation~\cite{alonso2020deep,gatesichapakorn2019ros}, and cinematographic staging~\cite{louarn2018automated}.
Their use in all these domains highlights the enduring significance of blueprints as a means of conveying abstract scene information for various purposes. 

In this paper, we present an approach for constructing a semantic editable blueprint from multiple viewpoints of a scene by leveraging prior semantic and depth information. 
Our novel \textbf{Bl}ueprint \textbf{N}eural \textbf{F}ield module, BluNF, generates a 2D semantic-aware editable blueprint that is robust to noise and sparse observations. 
Notably, BluNF enables user-friendly editing of NeRF representations. 
To the best of our knowledge, we are the first to employ an implicit neural field to construct a 2D semantic-aware blueprint for editing purposes, as most bird's-eye-view methods rely on CNNs or transformers~\cite{li2022bevformer}.

First, BluNF is capable of generating blueprints even in scenarios with incomplete depth or semantic information, without explicit geometric constraints. It improves robustness compared to multi-view-based methods~\cite{hartley2003multiple} (MVR) or straightforward NeRF orthogonal projections (see top-part Figure~\ref{fig:teaser}).
Second, BluNF introduces a novel and intuitive editing approach for NeRF. By providing a concise and informative blueprint, BluNF enables easy editing and manipulation, including changes in appearance and object removal, achieved through a simple \textit{click-and-change} mechanism, as depicted in bottom-part Figure~\ref{fig:teaser}.

Our contributions are:
(i) BluNF, a novel module that generates a blueprint of a scene through an implicit neural field without explicit geometric constraints or supervision;
(ii) an underlying representation that is robust to sparse and noisy observations, enabling reliable semantic layer identification and outperforming multi-view-based or classical NeRF methods; and
(iii) a combination of BluNF with NeRF representations of the same scene, enabling intuitive user manipulations on the generated 2D blueprint (masking, appearance, removal), and enabling direct view rendering with NeRF incorporating the edits.

\section{Related work}
\noindent \textbf{Implicit Neural Representation.}
Over the past few decades, implicit neural representations (INRs) have become popular for representing complex scenes due to their high efficiency, broad capacity, and diverse applications \cite{lombardi2019neural,mildenhall2021nerf,li2020end}. 
Their core idea is to exploit neural network representations to directly fit the target output by learning from input data without explicit parameterization. INRs have been successfully applied to various areas, including meshes~\cite{zhang2020path,nimier2020radiative}, voxels and point clouds~\cite{lombardi2019neural,jiang2020sdfdiff}, and light fields~\cite{shi2020learning}. Recently, NeRF~\cite{mildenhall2021nerf} has shown success in realistic viewpoint reconstructions, synthesizing photorealistic images of unseen views, while preserving geometric consistency and handling reflective lighting conditions.

To enhance the learning process and improve the representation performance of INRs, numerous techniques have been introduced: SIREN~\cite{sitzmann2020siren} proposes a periodic activation function to fit complex and generic natural signals and derivative information such as images, acoustics, and spatial data. \cite{tancik2020fourier} propose a frequency-based Positional Encoding (PE) to extract features for improving fitting performance, especially in high-frequency areas. HashGrid~\cite{muller2022instant} leverages hash encoding to address the ambiguity and smoothing issues, achieving gain in both fitting performance and training speed. In our work, we show that choosing the appropriate encoding scheme can have a substantial impact on the INR performance (Section~\ref{sub:ablation}).

\noindent \textbf{NeRF.}
The original proposal for NeRF was made by Mildenhall et al~\cite{mildenhall2021nerf}, with the main contribution of a Positional Encoding-supported MLP network to encode 5D spatial and view angle information. The pixel-wise multi-views image is constrained by computing the ray-based volumetric rendering. Subsequent improvements and derivatives have been proposed in order to enhance the system's image synthesis performance~\cite{barron2022mip, barron2021mip}, robustness to sparse views~\cite{niemeyer2022regnerf}, and support for dynamic scenes~\cite{pumarola2021d, park2021nerfies}, etc. Moreover, NeRF has become a powerful system capable of producing not only photorealistic rendering images, but also fitting various genres of applications. For instance, Zhi et al.~\cite{zhi2021place} propose adding a semantic head parallel to the RGB one for semantic digit rendering. A similar task is also investigated in~\cite{vora2021nesf}, where the main idea is to exploit the density field during the NeRF training. 

Recently,~\cite{yang2021learning} shows how to edit disentangled objects by constructing multiple NeRF models dedicated to each object and then combining them for scene rendering. This enables translating, rotating and scaling pre-defined objects in the scene. 
A similar concept is adapted in~\cite{fu2022panoptic,kundu2022panoptic,zhang2021editable} by combining composition functionality with dynamic or semantic scene representations for more detailed user manipulations. 
GAN-based NeRF systems~\cite{schwarz2020graf,niemeyer2021giraffe,chen2022sem2nerf} provide another fresh perspective on the task of generating NeRF, yet do not enable manipulating pre-trained NeRF. 
In our work, we use the proposed BluNF for manipulating and editing a pre-trained NeRF representation of a scene.

\noindent \textbf{Layout.}
Scene reconstruction is a longstanding problem in computer vision and graphics~\cite{choi2015robust, popov2020corenet, zollhofer2018state}, traditionally tackled with 3D geometry techniques such as SfM~\cite{Schonberger_2016_CVPR} and SLAM~\cite{whelan2015elasticfusion, wang2019densefusion}. However, with the recent development of deep neural networks (DNNs), the reconstruction problem has evolved to encompass a variety of applications: from explicit supervised reconstruction~\cite{tateno2017cnn,choy20163d} to 2D BEV (Bird-Eye-View) extraction for autonomous driving applications~\cite{li2022bevformer, 9681287}, and to INR-based reconstruction~\cite{chabra2020deep}.  
For reconstruction, various systems are proposed for different targets, scales~\cite{derksen2021shadow,xiangli2022bungeenerf,turki2022mega}, and usage scenarios, ranging from terrain~\cite{derksen2021shadow,xiangli2022bungeenerf,turki2022mega} and urban~\cite{tancik2022blocknerf} to indoor scenes~\cite{pavlakos2022one} or surface reconstruction applications~\cite{azinovic2022neural}. 
For editing, various systems allow for customizations, such as geometric editing or object removal~\cite{yang2021learning, yuan2022nerf}, color or style changing~\cite{huang2022stylizednerf,kuang2022palettenerf}, and generative manipulation~\cite{schwarz2020graf,niemeyer2021giraffe}. 

Nevertheless, most editing methods suffer from non-intuitive handling of 3D content (particularly in NeRF). This requires users to possess prior knowledge, to provide semantic masks for each training image~\cite{yang2021learning}, manually annotate object contours~\cite{liu2022nerf, mirzaei2022spin} or indicate a pre-trained latent code that lacks detailed manipulation~\cite{devries2021unconstrained, niemeyer2021giraffe, schwarz2020graf}, or could involve batch operations~\cite{huang2022stylizednerf, kuang2022palettenerf}.
Previous works~\cite{devries2021unconstrained,chen2022tensorf} partially mention the idea of 2D layout for NeRF scene representation. However, they only focus on acceleration or latent controlled generation and not specifically intended for editing or semantic understanding tasks. 
Instead, we propose a blueprint neural representation that enables intuitive and efficient manipulation of the layout when combined with NeRF.

\section{Method}
\begin{figure}
  \centering
    \includegraphics[width=\linewidth]{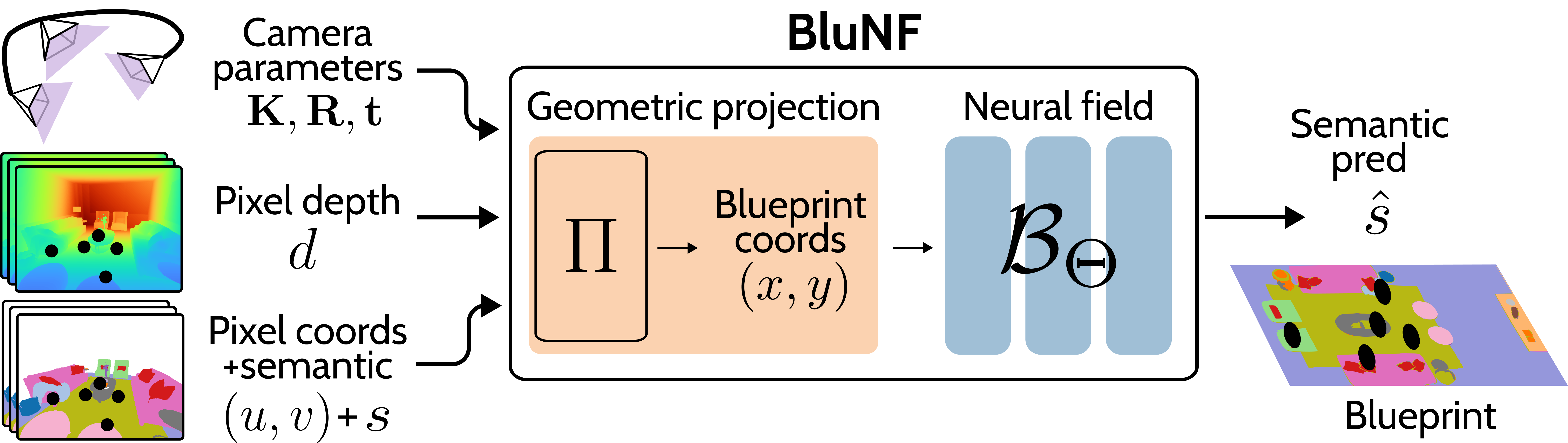}
    \caption{\textbf{Overview of BluNF training pipeline.} BluNF leverages camera parameters $\mathbf{K}, \mathbf{R}, \mathbf{t}$ and pixel depth $d$ 
    to establish a mapping from pixel coordinates $(u, v)$ to blueprint coordinates $(x, y)$. This mapping is achieved through the geometric projection module $\Pi$.
    The resulting blueprint coordinates $(x, y)$ are then fed as inputs to the neural field module $\mathcal{B}_\Theta$, which predicts the corresponding semantic label $\hat{s}$. We use the semantic label $s$ associated to pixel coordinates $(u, v)$ as supervision target.}
    \label{fig:main-train}
\end{figure}

In this work, we introduce the \textbf{Blu}eprint \textbf{N}eural \textbf{F}ield (BluNF) that generates a 2D semantic-aware blueprint of a 3D scene from sparse semantic views. 
For a given scene, a \textit{blueprint} refers to a top-view semantic floorplan, as illustrated in Figure~\ref{fig:teaser}.
In Section~\ref{sub:blunf_training}, we show that BluNF builds on neural implicit learning techniques by mapping pixel-wise semantic information to corresponding blueprint semantic labels constrained on a ray-based loss. 
In Section~\ref{sub:blunf_editing}, we elaborate on how BluNF 
improves the efficiency and reliability of scene editing tasks. 

\subsection{BluNF}
\label{sub:blunf_training}

\begin{figure}
  \centering
  \includegraphics[width=0.9\columnwidth]{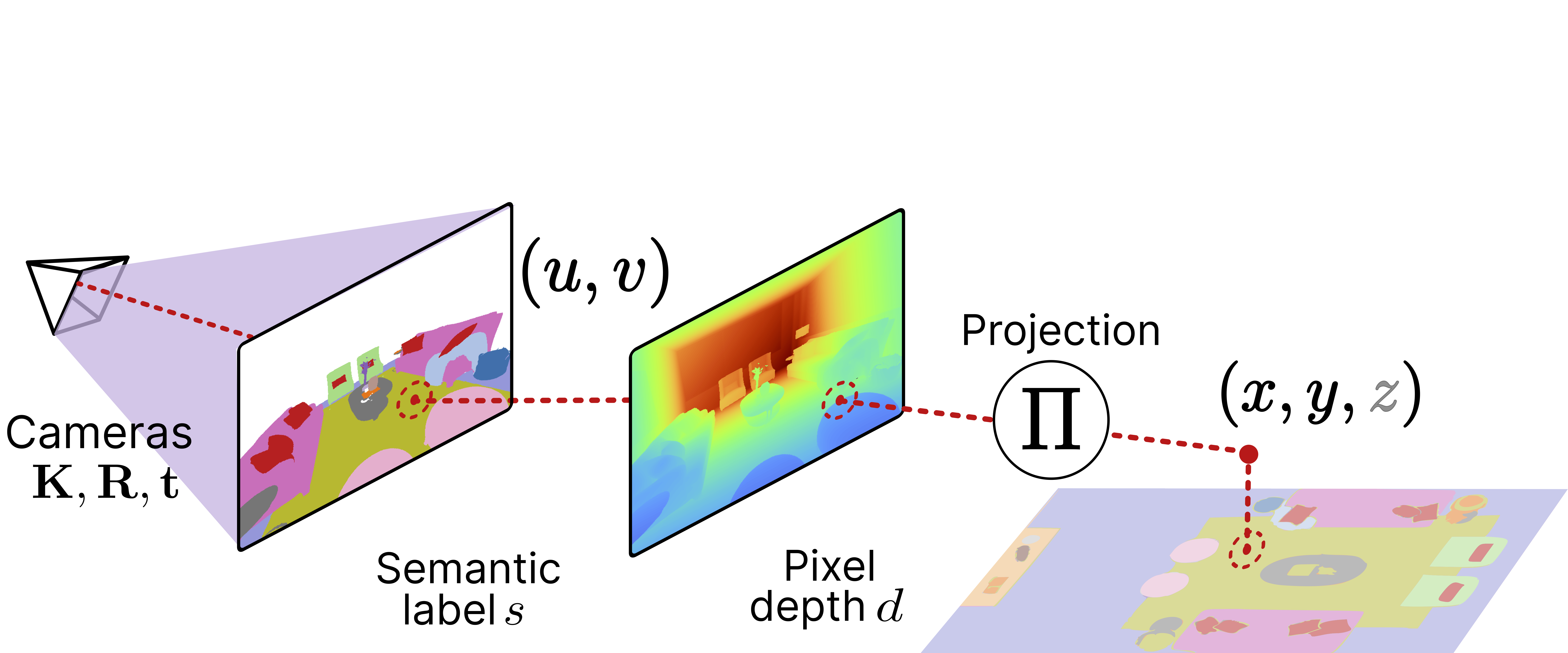}
  \caption{
  \textbf{View-to-blueprint projection.} 
  Given an image view captured by a camera with its intrinsic and extrinsic parameters $\mathbf{K}, \mathbf{R}, \mathbf{t}$, and a associated depth, the projection module $\mathbf{\Pi}$ projects one pixel's coordinates $(u, v)$, incorporating the associated depth $d$ back to 3D space $(x, y, z)$. We then project it onto the XY-plane (floor) to produce the 2D blueprint coordinates $(x, y)$.}
  \label{fig:projection}
\end{figure}

Figure~\ref{fig:main-train} provides an overview of the BluNF training pipeline, illustrating the two main modules: 
(i) a \textit{geometric projection} $\mathbf{\Pi}$ and (ii) a \textit{neural field} $\mathcal{B}_{\Theta}$.
To begin, we employ a \textit{geometric projection module} $\mathbf{\Pi}$ to map pixel coordinates $(u, v)$ from input views to 2D blueprint coordinates $(x, y)$.
Next, the \textit{neural field module} $\mathcal{B}_{\Theta}$ predicts a semantic label $\hat{s}$ from the input blueprint coordinates.
Thereby, BluNF effectively captures the underlying scene's semantic information, enabling to build an implicit representation of the blueprint. Below we detail these two modules.

\noindent \textbf{Projection module.}
The goal of the projection module is to transform the camera view coordinates $(u,v)$ to blueprint coordinates $(x,y)$. Figure~\ref{fig:projection} illustrates this process.
Given semantic views, we project each pixel coordinate $(u, v) \in \llbracket H, W \rrbracket$ along with its associated semantic label $s \in \llbracket 1, C_s \rrbracket$ onto the 2D semantic blueprint plan, where $C_s$ represents the number of semantic classes.
This projection, denoted as $\mathbf{\Pi}$, becomes feasible when the intrinsic matrix $\mathbf{K} \in \mathbb{R}^{3 \times 3}$ and extrinsic matrix $[\mathbf{R}|\mathbf{t}] \in \mathbb{R}^{3 \times 4}$ associated with the camera view are known, along with the depth.
By leveraging the projection matrix $\mathbf{\Pi}$ and the depth value $d \in \mathbb{R}$ at coordinates $(u, v)$, we  derive the 3D coordinates $(x, y, z) \in \mathbb{R}^3$:

\begin{equation} \label{eq:projection}
    \begin{bmatrix}
    x \\
    y \\
    z   \\
    1
    \end{bmatrix} = 
    \mathbf{\Pi}
    \begin{bmatrix}
    u.d \\
    v.d \\
    d   \\
    1
    \end{bmatrix}
    \text{, with }
    \mathbf{\Pi}^{-1} = \begin{bmatrix}
    \mathbf{K} & 0 \\
    0 & 1 \\
    \end{bmatrix} 
    \begin{bmatrix}
    \mathbf{R}  |  \mathbf{t}  \\
    \end{bmatrix} \quad ,
\end{equation}

Next, by projecting the 3D coordinates onto the floor -- i.e., the $XY$ plane --, we obtain the corresponding blueprint coordinates $(x, y) \in \mathbb{R}^2$. 
Consequently, for all pixel coordinates $(u, v)$ of scene views, we generate the inputs $(x, y)$ used as training data for the neural field module. 

\noindent \textbf{Neural field module.} 
The neural field module $\mathcal{B}_\Theta$ is designed to predict a semantic label $\hat{s}$ associated with the projected blueprint coordinates $(x, y)$.
$\mathcal{B}_\Theta$ consists of two submodules: (i) an encoding block that encodes blueprint coordinates $(x, y) \in \mathbb{R}^2$ into a higher-dimensional vector, and (ii) an MLP that maps the encoded input coordinates to a predicted semantic label $\hat{s}$. 
As depicted in Figure~\ref{fig:main-train}, after the projection of semantic labels from scene views, the neural field module learns an implicit representation of the scene's semantic blueprint.
We choose to rely on an implicit representation due to its robustness and the inherent ability to handle sparse inputs, as explained in prior works~\cite{mildenhall2021nerf,sitzmann2020siren}. 

\noindent \textbf{Training and loss.}  The BluNF pipeline is trained with a cross-entropy loss that compares the projected semantic label $s$ associated with the pixel coordinates $(u, v)$ and the predicted semantic label $\hat{s}$ generated by the neural field module. The parameters $\Theta$ of the MLP are then updated using standard gradient descent optimization. 

\noindent \textbf{Ambiguities.}  Geometric ambiguities may arise when projecting pixel coordinates from different views, resulting in the possibility of two different semantic labels being projected onto close blueprint coordinates. 
We incorporated the coordinate encoder in the neural field module to disentangle these geometric ambiguities, inspired by the encoding introduced in the NeRF paper~\cite{mildenhall2021nerf}.
By mixing both high- and low-frequency components, the encoder enables the MLP to distinguish points more accurately (high-frequency) while also smoothing out ambiguities (low-frequency).

\subsection{BluNF editing}
\label{sub:blunf_editing}

\begin{figure}
  \centering
    \includegraphics[width=\linewidth]{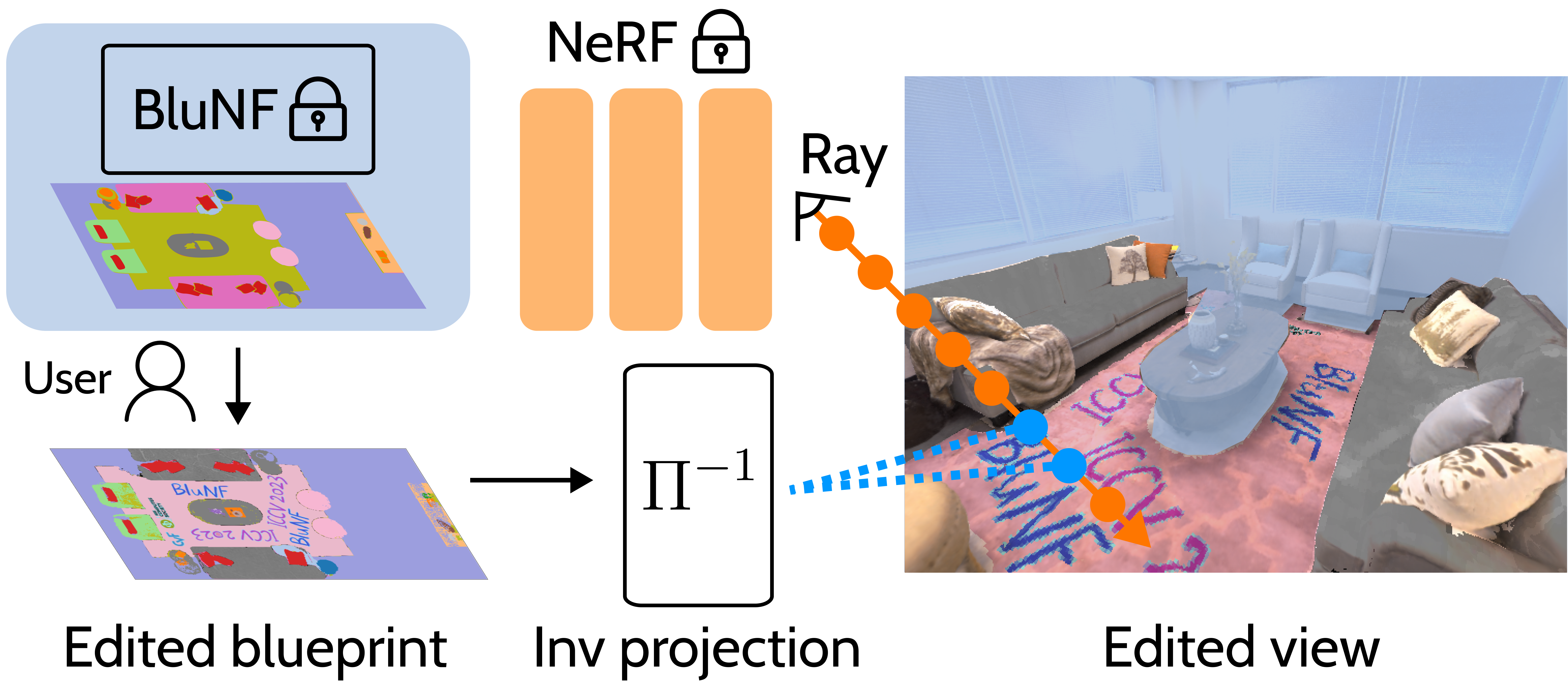}
    \caption{
    \textbf{Overview of BluNF editing pipeline.} It shows the process of leveraging a pre-trained BluNF model in conjunction with user interactions to edit a blueprint representation. During view rendering with a pre-trained NeRF, the NeRF samples (\textit{orange dots}) impacted by the edits are replaced by back-projected samples from the edited blueprint (\textit{blue dots}) using the inverse projection $\mathbf{\Pi^{-1}}$.}
    \label{fig:main-edit}
\end{figure}

Our BluNF representation is designed to enable user interactions on the blueprint, and when combined with the corresponding NeRF representation, it enables rendering any views with the edits. Figure~\ref{fig:main-edit} illustrates this process.
The blueprint consists of connected components, representing groups of pixels belonging to the same semantic class, forming distinct shapes.  Through user selection, these connected components can be modified by assigning new textures or completely removed from the 3D scene.
To incorporate the edits, we use the inverse of the projection $\mathbf{\Pi^{-1}}$, as defined in Section~\ref{sub:blunf_training}, to back-project the blueprint coordinates affected by the edits onto the 3D space, creating a set of 3D samples. During rendering, with a pre-trained NeRF model, we replace affected samples along the rays based on the back-projected edited samples. These sample replacements can involve adjustments to the color value, density value, or both.  For instance, removing an object entails cancelling the density value of the corresponding sample. 

\section{Experiments}
\noindent \textbf{Metrics.} 
To evaluate the performance of BluNF, we employ standard segmentation metrics~\cite{long2015fcn}: pixel accuracy (\textit{pAcc}) and frequency-weighted IoU (\textit{fwIoU}). Additionally, we report the completeness factor (\textit{comp.}) that measures the ratio of the valid output area to the total blueprint plan to assess the completeness of the reconstructed blueprint. To investigate the robustness of each method against partial observations, common in reconstruction problems~\cite{niemeyer2022regnerf, hartley2003multiple, Schonberger_2016_CVPR}, we report results for each dataset and metric with varying numbers of input frames $(90, 45, 9)$.

\noindent \textbf{Datasets.} 
We test on four scenes from two 3D indoor datasets: Replica~\cite{straub2019replica} - \textit{room 0} referred to as \textit{R1}, and \textit{room 2} referred to as \textit{R2} -, and MatterPort3D~\cite{Matterport3D} - \textit{gZ6f7yhEvPG} referred to as \textit{M1}, and \textit{pLe4wQe7qrG} referred to as \textit{M2}.
Replica is a synthetic dataset, while MatterPort3D contains real-world RGB-D information. For both datasets, we manually collect the ground truth blueprints from the given 3D models by computing the orthogonal projection with carefully hand-selected area and culled objects to avoid vertical occlusion (Figure~\ref{fig:main-comparison}). See supplementary for more details.

\subsection{Reconstruction comparison}
\label{sec:sota}
\begin{figure*}
  \centering
  \includegraphics[width=\textwidth]{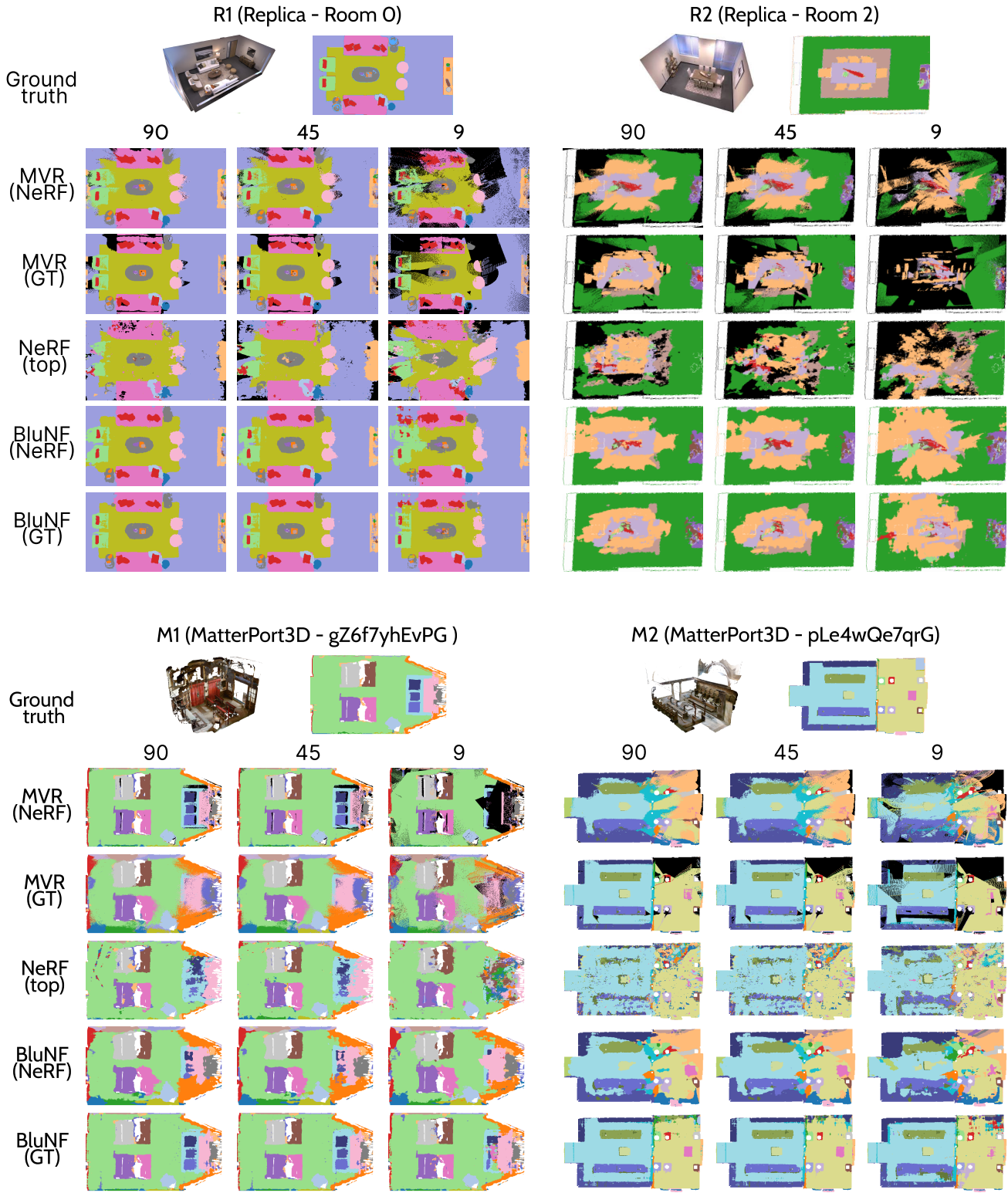}
  \caption{
  \textbf{Visualizations of generated blueprints for different methods} on the \textit{Replica - R1} (top-left), \textit{Replica - R2} (top-right), textit{MatterPort3D - M1} (bottom-left) and \textit{MatterPort3D - M2}  (bottom-right) datasets. Each row corresponds to a method: NeRF top-view, MVR with or without ground-truth depth, and BluNF with or without ground-truth depth. Each column displays results for a different number of input frames.
  }
  \label{fig:main-comparison}
\end{figure*}

\begin{table*}[h]
    \centering
    \begin{adjustbox}{minipage=\linewidth,scale=0.85}
    \begin{subtable}[h]{\linewidth}
    \centering
    \begin{tabular}{@{}ll@{\hspace{3pt}}l@{\hspace{3pt}}l@{\hspace{7pt}}l@{\hspace{3pt}}l@{\hspace{3pt}}l@{\hspace{7pt}}l@{\hspace{3pt}}l@{\hspace{3pt}}l@{\hspace{3pt}}ll@{\hspace{3pt}}l@{\hspace{3pt}}l@{\hspace{7pt}}l@{\hspace{3pt}}l@{\hspace{3pt}}l@{\hspace{7pt}}l@{\hspace{3pt}}l@{\hspace{3pt}}l@{}}
    \toprule
    \multirow{2}{*}{\textbf{Method (depth)}} & \multicolumn{9}{c}{\cellcolor{red!25}{\textbf{R1 (room 0)}}} & & \multicolumn{9}{c}{\cellcolor{red!25}{\textbf{R2 (room 2)}}}                                                  \\
    & \multicolumn{3}{c}{\cellcolor{gray!15}{pACC $\uparrow$}} & \multicolumn{3}{c}{\cellcolor{gray!15}{fwIoU $\uparrow$}} & \multicolumn{3}{c}{\cellcolor{gray!15}{comp. $\uparrow$}} & & \multicolumn{3}{c}{\cellcolor{gray!15}{pACC $\uparrow$}} & \multicolumn{3}{c}{\cellcolor{gray!15}{fwIoU $\uparrow$}} & \multicolumn{3}{c}{\cellcolor{gray!15}{comp.  $\uparrow$}} \\ 
    \textbf{$\#$ frames}                      & $90$ & $45$ & $9$ & $90$ & $45$ & $9$ & $90$ & $45$ & $9$ & & $90$ & $45$ & $9$ & $90$ & $45$ & $9$ & $90$ & $45$ & $9$   \\
    \midrule
    
    
    \textbf{MVR (GT)} & $88.1$ & $87.0$ & $65.4$ & $82.7$  & $81.7$ & $61.3$ & $95.2$    & $94.1$    & $72.2$  & & $60.6$ & $56.6$ & $31.1$ & $53.2$ & $50.3$ & $29.5$ & $51.8$ & $47.9$ & $26.1$    \\

    \textbf{BluNF (GT)} & $\mathbf{92.6}$ & $\mathbf{92.7}$ & $\mathbf{90.0}$ & $\mathbf{86.7}$ & $\mathbf{86.9}$ & $\mathbf{82.3}$ & $\mathbf{100}$ & $\mathbf{100}$ & $\mathbf{100}$ & & $\mathbf{71.9}$ & $\mathbf{71.9}$ & $\mathbf{58.6}$ & $\mathbf{60.9}$ & $\mathbf{60.9}$ & $\mathbf{49.6}$ & $\mathbf{100}$ & $\mathbf{100}$ & $\mathbf{100}$ \\

    \midrule
    \textbf{NeRF (top-view)} & $82.2$ & $80.7$ & $69.2$ & $72.6$  & $71.4$ & $58.4$ & $97.7$    & $97.3$    & $93.1$ &  & $51.5$ & $40.4$ & $31.2$ & $44.8$ & $34.6$ & $27.3$ & $56.6$ & $49.3$ & $47.4$ \\ 
    
    \textbf{MVR (NeRF)} & $88.3$ & $86.8$ & $63.1$ & $81.0$  & $79.5$ & $57.2$ & $99.1$    & $97.8$    & $74.3$  & & $46.9$ & $44.3$ & $26.0$ & $38.7$ & $36.4$ & $22.3$ & $48.4$ & $46.2$ & $31.4$ \\
    
    \textbf{BluNF (NeRF)} & $\mathbf{90.6}$ & $\mathbf{90.5}$ & $\mathbf{85.5}$ & $\mathbf{83.6}$ & $\mathbf{83.6}$ & $\mathbf{76.6}$ & $\mathbf{100}$ & $\mathbf{100}$ & $\mathbf{100}$ & & $\mathbf{62.1}$ & $\mathbf{63.9}$ & $\mathbf{50.6}$ & $\mathbf{51.9}$ & $\mathbf{52.9}$ & $\mathbf{41.8}$ & $\mathbf{100}$ & $\mathbf{100}$ & $\mathbf{100}$ \\ 
    
    \bottomrule
    \end{tabular}
    \caption{\textbf{Replica dataset}.}
    \label{tab:main-comparison-replica}
    \vspace{0.3cm}
  \end{subtable}
  \hfill
  \begin{subtable}[h]{\linewidth}
    \centering
    \begin{tabular}{@{}ll@{\hspace{3pt}}l@{\hspace{3pt}}l@{\hspace{7pt}}l@{\hspace{3pt}}l@{\hspace{3pt}}l@{\hspace{7pt}}l@{\hspace{3pt}}l@{\hspace{3pt}}l@{\hspace{3pt}}ll@{\hspace{3pt}}l@{\hspace{3pt}}l@{\hspace{7pt}}l@{\hspace{3pt}}l@{\hspace{3pt}}l@{\hspace{7pt}}l@{\hspace{3pt}}l@{\hspace{3pt}}l@{}}
    \toprule
    \multirow{2}{*}{\textbf{Method (depth)}} & \multicolumn{9}{c}{\cellcolor{blue!25}{\textbf{M1 (gZ6f7yhEvPG)}}} & & \multicolumn{9}{c}{\cellcolor{blue!25}{\textbf{M2 (pLe4wQe7qrG)}}}                                                  \\
    & \multicolumn{3}{c}{\cellcolor{gray!15}{pACC $\uparrow$}} & \multicolumn{3}{c}{\cellcolor{gray!15}{fwIoU $\uparrow$}} & \multicolumn{3}{c}{\cellcolor{gray!15}{comp. $\uparrow$}} & & \multicolumn{3}{c}{\cellcolor{gray!15}{pACC $\uparrow$}} & \multicolumn{3}{c}{\cellcolor{gray!15}{fwIoU $\uparrow$}} & \multicolumn{3}{c}{\cellcolor{gray!15}{comp.  $\uparrow$}} \\ 
    \textbf{\# frames}                      & $90$ & $45$ & $9$ & $90$ & $45$ & $9$ & $90$ & $45$ & $9$ &  & $90$ & $45$ & $9$ & $90$ & $45$ & $9$ & $90$ & $45$ & $9$   \\
    \midrule
    
    
    \textbf{MVR (GT)} & $87.8$ & $86.8$ & $73.1$ & $82.8$  & $81.9$ & $69.4$ & $90.3$    & $89.1$    & $74.7$  &  & $81.5$ & $80.3$ & $65.5$ & $76.5$ & $75.3$ & $61.6$ & $77.4$ & $76.5$ & $64.4$ \\
    
    \textbf{BluNF (GT)} & $\mathbf{89.0}$ & $\mathbf{89.1}$ & $\mathbf{84.2}$ & $\mathbf{82.9}$ & $\mathbf{82.9}$ & $\mathbf{76.5}$ & $\mathbf{100}$ & $\mathbf{100}$ & $\mathbf{100}$ & & $\mathbf{ 87.4}$ & $\mathbf{87.6}$ & $\mathbf{82.8}$ & $\mathbf{80.7}$ & $\mathbf{80.9}$ & $\mathbf{75.5}$ & $\mathbf{100}$ & $\mathbf{100}$ & $\mathbf{100}$ \\
    
    \midrule
    \textbf{NeRF (top-view)} & $\mathbf{80.3}$ & $\mathbf{80.0}$ & $\mathbf{73.1}$ & $\mathbf{73.0}$  & $\mathbf{72.0}$ & $\mathbf{64.9}$ & $\mathbf{100}$   & $\mathbf{100}$   & $\mathbf{100}$ & & $\mathbf{62.5}$ & $\mathbf{62.8}$ & $\mathbf{59.5}$ & $49.0$ & $49.5$ & $\mathbf{47.4}$ & $\mathbf{100}$ & $\mathbf{100}$ & $\mathbf{100}$ \\
    
    \textbf{MVR (NeRF)} & $62.0$ & $61.3$ & $54.5$ & $54.5$  & $53.9$ & $47.7$ & $95.5$    &$95.2$    & $90.5$ & & $52.3$ & $52.5$ & $45.9$ & $42.9$ & $43.1$ & $36.9$ & $87.3$ & $86.8$ & $76.9$ \\
    
    \textbf{BluNF (NeRF)} & $67.9$ & $67.9$ & $64.2$ & $59.0$ & $59.4$ & $56.5$ & $\mathbf{100}$ & $\mathbf{100}$ & $\mathbf{100}$ & & $62.0$ & $62.3$ & $57.3$ & $\mathbf{53.0}$ & $\mathbf{53.1}$ & $47.2$ & $\mathbf{100}$ & $\mathbf{100}$ & $\mathbf{100}$ \\ 
    
    \bottomrule
    \end{tabular}
    \caption{\textbf{MatterPort3D dataset}.}
    \label{tab:main-comparison-matterport}
    
\end{subtable}
\end{adjustbox}

\caption{
\textbf{Comparison to the state of the art on \textit{Replica} (a) and \textit{MatterPort3D} (b) for varying number of frames (90, 45,9).} 
For both subtables, we report results for 2 scenes (left and right), the top part corresponds to results with ground-truth depth (GT), and the bottom part without, i.e., it uses the estimated-by-NeRF depth.}
\label{tab:main-comparison}
\end{table*} 

\noindent \textbf{Compared methods.}
We compare BluNF to two other methods:
First, the multi-view reconstruction (MVR) method that directly projects and gathers all pixels in the dataset to generate the blueprint. 
Second, the \textit{NeRF (top)} method that computes the blueprint by sampling orthogonal rays vertically to the floor in a pre-trained NeRF under the same input images. \textit{NeRF (top)} is sensitive to the height selection:
too low height leads to over-trimmed objects; too high height risks of being influenced by in-the-air or corner artifacts which are less supervised. For a fair comparison, we report the best fwIoU result among different sampled heights. For BluNF, we report results with two input depth sources: the ground-truth (\textit{GT}) and the NeRF-estimated depth (\textit{NeRF}). Additionally, for all methods, we use the semantic maps provided in the datasets.

\noindent \textbf{Quantitative results.}
We compare our proposed BluNF to both methods mentioned above and report the results in Table~\ref{tab:main-comparison}. Overall, we observe that for the four scenes, BluNF outperforms the other methods for all metrics and all options, i.e., number of input frames and nature of depth.

More precisely, for \textbf{Replica scenes}: (i) \textit{R1} (subtable top-left): BluNF performs the best for both depth options: for instance,  with $90$ frames as input, it reaches pACC of $92.6\%$ and $90.6\%$ with and without ground truth depth, respectively, against $88.1\%$  and  $88.3\%$ for MVR. 
(ii) \textit{R2} (subtable top-right): MVR and BluNF using the ground-truth depth achieve relatively high scores for all metrics and all numbers of frames. However, BluNF generally outperforms MVR, achieving higher scores for pACC, fwIoU, and completeness across all numbers of frames. For example, using $90$ frames, BluNF achieves pACC, fwIoU, and completeness scores of $71.9\%$, $60.9\%$, and $100\%$, respectively, while MVR achieves scores of $60.6\%$, $53.2\%$, and $51.8\%$, respectively.
For \textbf{MatterPort3D scenes}: (i) \textit{M1} (subtable bottom-left): with ground-truth depth (GT), BluNF is also better than MVR for both pACC and fwIoU; for instance, at $90$ frames it reaches $89.0\%$ and $82.9\%$ against $87.8\%$ and $82.8\%$, for both metrics respectively.
When using NeRF-estimated depth, even though BluNF  outperforms MVR by $+9.7\%$ in pACC and $+8.8\%$ in fwIoU for $9$ frames; \textit{NeRF (top)} performs the best with $73.1\%$ of pACC and $64.9\%$ for $9$ frames as well. 
(i) \textit{M2} (subtable bottom-right): with ground-truth depth, the results are similar, with BluNF again outperforming MVR for all metrics and numbers of frames. 
Interestingly, the scores are in most cases lower for MatterPort3D scenes than for Replica scenes. We attribute this discrepancy to the intrinsic quality differences between the datasets. The MatterPort3D datasets rely on RGB-D acquisitions that vary in quality, resulting in a relatively poorer quality overall. For a more detailed discussion on this matter, please refer to the supplementary material.

\noindent \textbf{Qualitative results.}
Figure~\ref{fig:main-comparison} shows visual results of all methods for both datasets. It highlights the failure of other methods, notably MVR, and especially when the number of frames is low (blacks areas in blueprints in rows 1, 2, 6, 7; columns 3, 6). Moreover, we note the high quality of BluNF-generated blueprint (row 5, 10; columns 1, 4), in contrast to the others (rows 2, 3, 7, 8; columns 1, 4).

\subsection{Ablation study}
\label{sub:ablation}

\begin{figure}
    \centering
    \scalebox{0.29}{\input{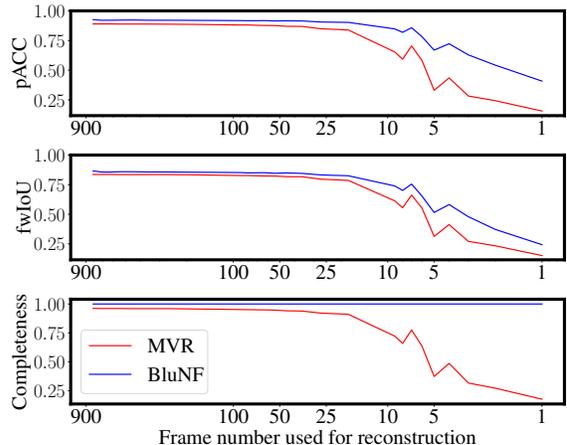}}
    \caption{
    \textbf{Ablation of number of input frames for BluNF and MVR on Replica-R1~\cite{straub2019replica}. } BluNF outperforms MVR in all metrics for all input frame numbers. Their gap is more notable when reducing to few input frames, i.e., less than 5.
    }
    \label{fig:ablation_fn}
\end{figure}

In this section, we present the results of four ablation studies to highlight the influence of: 
(i) the number of frames; 
(ii) the impact of depth;
(iii) the analysis of completeness;
(iv) the different types of encoding.
Most discussions and ablations are presented with the Replica R1 dataset, unless stated otherwise.

\noindent \textbf{Number of Frames.}
The number of frames is a critical factor for almost all reconstruction methods, including both geometric-based and learning-based approaches~\cite{niemeyer2022regnerf,Schonberger_2016_CVPR}. A low overlapping of multi-view information can lead to missing parts or erroneous estimations. 
In Figure~\ref{fig:ablation_fn}, we compare BluNF against MVR for different number of frames gradually reduced from 900 to 1 for Replica \textit{R1}.  We observe that our BluNF outperforms MVR for all metrics. Specifically, when the number of frames is high (e.g. some hundreds), both methods perform similarly for all metrics. However, the more the number of frames reduces, the more the gap between the two methods increases. 
This gap becomes even more pronounced when the number of frames is very low (e.g., fewer than 5 frames). 
This highlights the fact that leveraging the filling capacity of implicit representations used by BluNF leads to significantly better performance. 
We also observe this trend in Table~\ref{tab:main-comparison-replica}.

\noindent \textbf{Impact of depth.}
Here, we examine the impact of depth on our proposed BluNF by comparing the performance of BluNF with GT and NeRF-based depths, i.e., rows 2 and 5 in Table~\ref{tab:main-comparison-replica} and~\ref{tab:main-comparison-matterport}. 
For \textbf{Replica dataset \textit{R1}}, the nature of depth (GT or NeRF-estimated) does not influence much the performances of BluNF. Between ground-truth and estimated depth, BluNF drop around $5\%$ of pACC and fwIoU. 
For \textbf{MatterPort3D \textit{M1}}, the impact of depth is more prominent: the fwIoU of BluNF with 45 frames decreases from $82.9\%$ with ground truth depth to $59.4\%$ without.
Those two behaviours can be explained by the contrasted quality between the synthetic and real-acquired depth information. Refer to the supplementary material for more details.

\noindent \textbf{Analysis of completeness.}
Regarding completeness, BluNF exploits a continuous neural field and therefore its completeness is always maximum ($100\%$).
Nevertheless, note that for MVR this is an important drawback, and it gets worse when the number of frames decreases. For Replica \textit{R1} in Table~\ref{tab:main-comparison-replica}, \textit{MVR (GT)} completeness decreases from $95.2\%$ with $90$ frames, to $72.2\%$ with $9$ frames. 
In addition, it is interesting to see that NeRF-estimated depth increases the completeness of MVR. For instance, on MatterPort3D \textit{M1} in Table~\ref{tab:main-comparison-matterport} with $9$ frames, it increases from $74.7\%$ with ground truth, to $90.5\%$ without. It is explained by the error on the estimated depth, making the projection fuzzy, yet less accurate.
Note that \textit{NeRF (top)} achieves full completeness on MatterPort3D as there is no ceiling semantic class, on the contrary of Replica, polluted by ceiling predictions.

\begin{table}[]
\centering
\resizebox{\columnwidth}{!}{
\begin{tabular}{@{}l@{\hspace{3pt}}l@{\hspace{3pt}}l@{\hspace{3pt}}l@{\hspace{10pt}}l@{\hspace{3pt}}l@{\hspace{3pt}}l@{}}
\toprule
\textbf{Encoding type} & \multicolumn{3}{c}{\cellcolor{gray!15}{pACC $\uparrow$}} & \multicolumn{3}{c}{\cellcolor{gray!15}{fwIoU $\uparrow$}} \\
\textbf{Number of frames}                        & $90$    & $45$     & $9$      & $90$     & $45$     & $9$      \\ \midrule
\textbf{No encoding}                                 & $85.4$ & $85.5$ & $73.3$ & $77.0$  & $76.2$ & $63.0$ \\
\textbf{Hash}                               & $91.8$ & $91.6$ & $84.8$ & $85.2$  & $84.9$ & $73.9$ \\ 
\textbf{SIREN}                              & $91.0$ & $91.0$ & $84.5$ & $84.1$  & $84.2$ & $75.2$ \\ 
\textbf{PE}                                 & $\mathbf{92.6}$ & $\mathbf{92.7}$ & $\mathbf{90.0}$ & $\mathbf{86.7}$  & $\mathbf{86.9}$ & $\mathbf{82.3}$ \\
\bottomrule
\end{tabular}
}
\caption{
\textbf{Ablations of coordinate encoding} in BluNF on Replica \textit{R1} with ground-truth depth. 
}
\label{tab:module-ablation}
\end{table}

\begin{figure}
  \centering
  \begin{subfigure}[b]{0.23\textwidth}
    \centering
    \includegraphics[width=\textwidth]{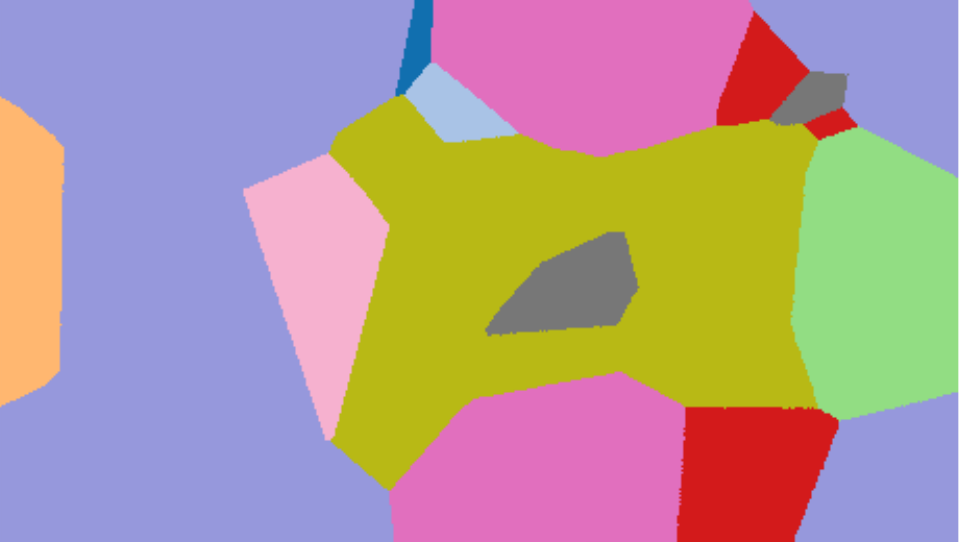}
    \caption{w/o PE}
    \label{fig:no-pe}
  \end{subfigure}\hfill
  \begin{subfigure}[b]{0.23\textwidth}
    \centering
    \includegraphics[width=\textwidth]{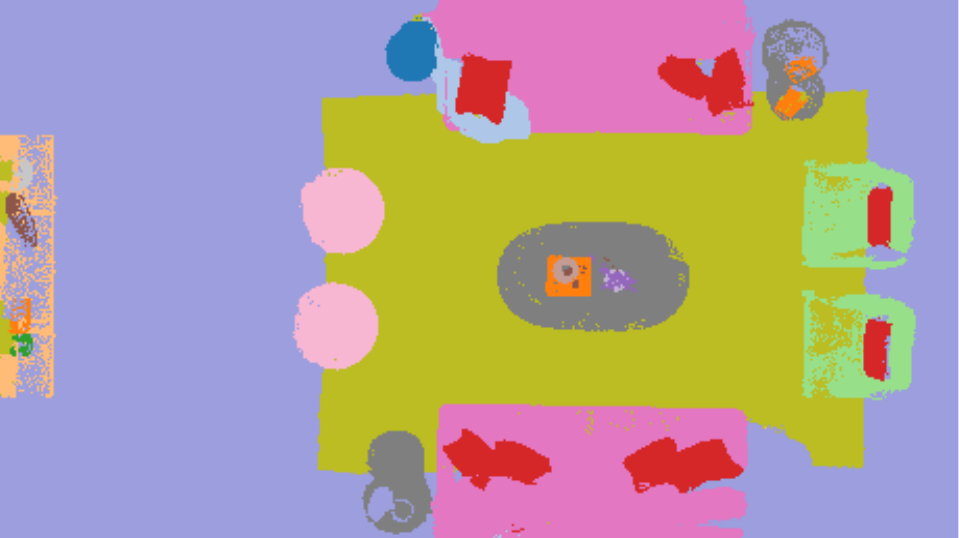}
    \caption{w/ PE}
    \label{fig:pe}
  \end{subfigure}
  \caption{\textbf{Visual comparison of different input encodings.} (a) Result with positional encoding (PE), and (b) result without positional encoding (w/o PE).}
  \label{fig:pe-ambiguity}
\end{figure}

\noindent \textbf{BluNF input encoding.}
We compare in Table~\ref{tab:module-ablation} the effectiveness of different types of encoding: (i) no encoding; (ii) the hash encoding from Instant-NGP~\cite{muller2022instant}; (iii) a SIREN-based BluNF~\cite{sitzmann2020siren}; and (iv) the standard positional encoding  (PE) introduced in NeRF~\cite{mildenhall2021nerf}.
Firstly, the results validate our claim in Section~\ref{sub:blunf_training} that encoding helps to disentangle the geometric ambiguities associated with the projection module. Specifically, there is a significant gap of up to $19.6\%$ in fwIoU between with and without PE encoding schemes when using 9 input frames.
In addition, these results show that with a high number of frames, the encoding type does not affect the results; the difference of pACC and fwIoU between the different encoding is around $\pm 1\%$. Nevertheless, when the number of frames decreases, \textit{Hash} and \textit{SIREN} are highly impacted. For instance for $9$ frames, we report a pACC of $84.8\%$ and $84.5\%$ and a fwIoU of $73.9\%$ and $75.2\%$ respectively. In contrast \textit{PE} seems much less impacted with an pACC of $90.0\%$ and a fwIoU of $82.3\%$.
Figure~\ref{fig:pe-ambiguity} presents a visual comparison of the influence of input encoding on the resulting blueprints. We compare blueprints generated with and without positional encoding. The comparison provides evidence supporting our claim from Section~\ref{sub:blunf_training} that positional encoding helps disentangle projection ambiguities. Blueprint with PE exhibits enhanced detail and accuracy, while those without encoding lack detail and are dominated by ambiguous shapes.

\subsection{Editing comparison}
\label{sub:editing}

\begin{figure}
  \centering
  \includegraphics[width=.95\columnwidth]{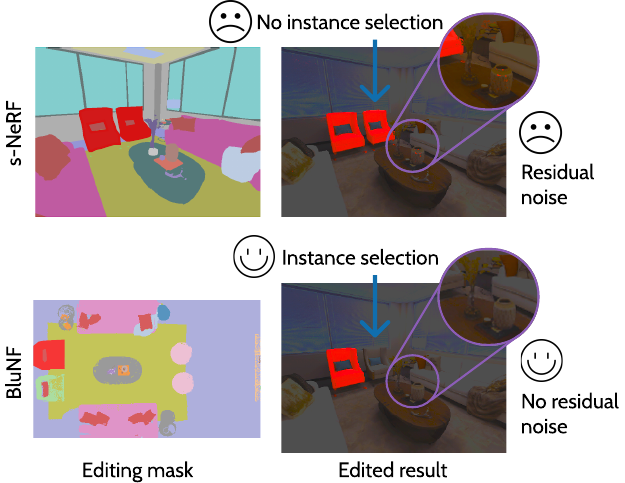}
  \caption{\textbf{Editing result comparison.} (Top) Semantic-NeRF-based editing results and (bottom) our proposed BluNF editing result.}
  \label{fig:edit-comparison}
\end{figure}

\noindent \textbf{Comparison to baselines.} 
In this work, we compare our BluNF editing pipeline with two baselines based on semantic-NeRF~\cite{zhi2021place}. 
The first baseline involves the user choosing a semantic class to edit for the entire scene and using the rendered semantic view of semantic-NeRF as a mask to apply edits on the RGB view. However, as shown in Figure~\ref{fig:edit-comparison}, this approach has several drawbacks compared to our proposed method BluNF. Specifically, as demonstrated by the blue arrow the two armchairs are selected due to belonging to the same semantic class. In contrast, BluNF enables us to select specific instances within the same semantic class. 
Additionally, the purple zoom illustrates that using every pixel of the same semantic class leads to artifacts intrinsic to semantic-NeRF, while BluNF, relying on connected components, avoids selecting these artifacts, highlighting the flexibility of BluNF.
The second baseline involves using the same mask selection as in BluNF, requiring the user to select connected components for each semantic view. However, this approach is impractical as it would demand the user to manually select connected components for all frames, emphasizing the generalization power of BluNF.

\begin{figure}
  \centering
  \includegraphics[width=.95\columnwidth]{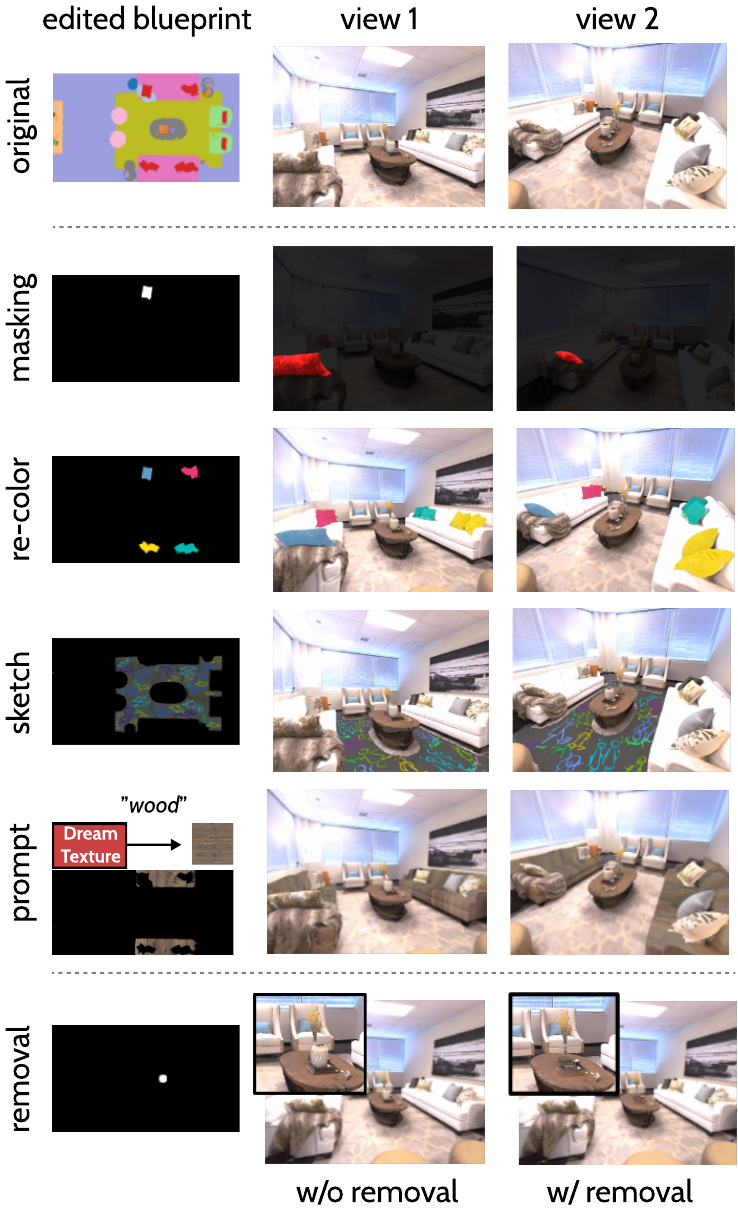}
  \caption{
  \textbf{Show cases of applications enabled by BluNF.} Examples of masking, recoloring, prompting, hand-sketching and instance removal.}
  \label{fig:edit-results}
\end{figure}

\noindent \textbf{Qualitative results.} Here we present the results of the BluNF editing pipeline introduced in Section~\ref{sub:blunf_editing}. Figure~\ref{fig:edit-results} illustrates the outcomes, where the first column shows the selected blueprint area for each application, while the second and third columns illustrate two different synthesized views. 
The applications we showcase are \textbf{Object selection/masking}: Our system enables the user to select instances for editing by clicking on the generated blueprint. Grouped instance is able to back-project to 3D as a high-quality dynamic mask; \textbf{Object recoloring}: The user can change the color of the selected area by simply applying color blending; \textbf{Prompt Re-texturing}: The user can generate a 2D texture from a text-prompt with an external model~\cite{dreamtextures}. Our system supports direct 2D textured mapping on the blueprint, which can be printed on the surface of selected objects; \textbf{Sketching}: Our system supports sketching or painting for editing instances; \textbf{Instance Removal}: Our system supports primitive instance removal by setting the density of the selected area to zero without requiring re-training of the NeRF. In the last row, we show that we remove a vase from the table by selecting it on the blueprint.

\section{Conclusion}
In this work, we introduced BluNF, a novel module capable of generating 2D semantic-aware editable blueprints of scenes. By leveraging implicit learning with semantic and depth priors, BluNF demonstrates superior performance compared to geometric and semantic NeRF-based methods. 
BluNF combined with NeRF allows various intuitive editing applications. It bridges the gap left by standard editing methods, limited by their lack of 3D spatial understanding when relying on traditional 2D interaction tools.

\section{Acknowledgment}
We would like to thank Nicolas Dufour for proofreading and the anonymous reviewers for their feedback. 
This work was supported by ANR-22-CE23-0007 project and the Hi!Paris collaborative project and scholarship.

{\small
\bibliographystyle{ieee_fullname}
\bibliography{short,main}
}

\end{document}